\documentclass{article}

\usepackage{PRIMEarxiv}

\usepackage[utf8]{inputenc} % allow utf-8 input
\usepackage[T1]{fontenc}    % use 8-bit T1 fonts
\usepackage{hyperref}       % hyperlinks
\usepackage{url}            % simple URL typesetting
\usepackage{booktabs}       % professional-quality tables
\usepackage{amsfonts}       % blackboard math symbols
\usepackage{nicefrac}       % compact symbols for 1/2, etc.
\usepackage{microtype}      % microtypography
\usepackage{lipsum}
\usepackage{fancyhdr}       % header
\usepackage{graphicx}       % graphics
\graphicspath{{media/}}     % organize your images and other figures under media/ folder
\usepackage{multirow}
\usepackage{subcaption}

\usepackage[acronym]{glossaries}

\makeglossaries
\newacronym{LAN}{LAN}{Local Area Networks}

\newacronym{saas}{SaaS}{Software as Service}

\newacronym{nlp}{NLP}{Natural Language Processing}

\newacronym{OSI}{OSI}{Open Systems Interconnection}

\newacronym{POC}{POC}{Proof of Concept}

\newacronym{FRS}{FRS}{Face Recognition Systems}

\newacronym{PRISMA}{PRISMA}{Preferred Reporting Items for Systematic Reviews and Meta-Analyses}

\newacronym{bmm}{BMM}{Bernoulli Mixture Model}

\newacronym{fvbmm}{FV-BMM}{Fisher Vectors - Bernoulli Mixture Model}

\newacronym{GMM}{GMM}{Gaussian Mixture Model}

\newacronym{hmm}{HMM}{Hidden Markov Model}

\newacronym{SVM}{SVM}{Support Vector Machine}

\newacronym{IoC}{IoC}{Indicator of Compromise}	

\newacronym{3dhog}{3DHOG}{3-Dimensional Histogram of Gradients}	

\newacronym{hof}{HOF}{Histogram of Optical Flow}

\newacronym{GAN}{GAN}{Generative Adversarial Network}

\newacronym{SPN}{SPN}{Sensor Pattern Noise}

\newacronym{PCA}{PCA}{Principal Component Analysis}

\newacronym{surf}{SURF}{Speeded Up Robust Features}

\newacronym{esurf}{eSURF}{3D extended Speeded Up Robust Features}

\newacronym{mips}{MIPs}{Motion Interchanges Patterns}

\newacronym{aic}{AIC}{Akaike Information Criterion}

\newacronym{mle}{MLE}{Maximum Likelihood Estimation}

\newacronym{fsmc}{FSMC}{Finite-State Markov Chain}

\newacronym{roi}{ROI}{Region of Interest}

\newacronym{ROC}{ROC}{Receiver Operating Characteristics}

\newacronym{eer}{EER}{Equal Error Rate}

\newacronym{auc}{AUC}{Area Under the Curve}

\newacronym{bof}{BoF}{Bag of Features}

\newacronym{fps}{FPS}{Frames Per Second}

\newacronym{EFDF}{EFDF}{Early Fusion Deep Features}

\newacronym{FC}{FC}{Fully Connected}

\newacronym{ccr}{CCR}{Correct Classification Rate}

\newacronym{OCR}{OCR}{Optical Chacter Recognition}

\newacronym{bdt}{BDT}{Binary Dense Trajectories}

\newacronym{bidt}{BIDT}{Binary Improved Dense Trajectories}

\newacronym{mbh}{MBH}{Motion Boundaries Histograms}

\newacronym{ARA}{ARA}{Architecture Reference Architecture}

\newacronym{em}{EM}{Expectation Maximization}

\newacronym{CDF}{CDF}{Cumulative Distribution Function}

\newacronym{C3D}{C3D \cite{DTran15}}{Convolution 3D}

\newacronym{HMP}{HMP \cite{JAlmeida11}}{Histogram of Motion Patterns}

\newacronym{AE}{AE}{Auto Encoder}

\newacronym{SVR}{SVR}{Support Vector Regression}

\newacronym{LBP}{LBP \cite{MHeikkil09}}{Local Binary Patterns}

\newacronym{ResNet}{ResNet \cite{KHe16}}{Residual Network}

\newacronym{3DResNet}{3DResNet \cite{KHara18}}{3-Dimensional Residual Network}

\newacronym{DenseNet}{DenseNet \cite{GHuang17}}{Dense Network}

\newacronym{ENet}{ENet}{Elastic Network}

\newacronym{CNN}{CNN}{Convolutional Neural Network}

\newacronym{BERT}{BERT}{Bidirectional Encoder Representations from Transformers}

\newacronym{HOG}{HOG \cite{NDalal05}}{Histogram of Oriented Gradients}

\newacronym{IV3}{IV3 \cite{CSzegedy16} }{Inception Version 3}

\newacronym{LSTM}{LSTM}{Long Short Term Memory}

\newacronym{GloVe}{GloVe \cite{JPennington14}}{Global Vectors}

\newacronym{FL}{FL}{Fuzzy Logic}

\newacronym{TSN}{TSN \cite{LWang16}}{Temporal Segment Networks}

\newacronym{DNN}{DNN}{Deep Neural Network}

\newacronym{FV}{FVs}{Fisher Vectors}

\newacronym{MLP}{MLP}{Multi-Layer Perceptron}

\newacronym{kNN}{kNN}{k-Nearest Neighbour}

\newacronym{AMNet}{AMNet \cite{Fajtl18}}{Attention-based Memorability estimation Network}

\newacronym{LASSO}{LASSO \cite{RTibshirani96}}{Least Absolute Shrinkage and Selection Operator}

\newacronym{I3D}{I3D \cite{Carreira17}}{Two-Stream Inflated 3D ConvNet}

\newacronym{GRU}{GRU \cite{KCho14}}{Gate Recurrent Unit}    

\newacronym{FFT}{FFT}{Fast Fourier Transform}    

\newacronym{NIST}{NIST}{National Institute of Standards and Technology}    

\newacronym{FRSLLS}{FRSLLS}{Face Research Lab London Set}

\newacronym{SOTA}{SOTA}{state-of-the-art}

\newacronym{MAP}{MAP}{Maximum A posterior Probability}

\newacronym{mAp}{mAp}{mean Average precision}

\newacronym{VGG}{VGG}{Visual Geometry Group}

\newacronym{FFHQ}{FFHQ}{Flick Faces High Quality}

\newacronym{DCT}{DCT}{Discreet Cosine Transform}

\newacronym{ELBO}{ELBO}{Evidence LOwer BOund}

\newacronym{CELEBA-HQ}{CELEBA-HQ}{Celebrities A High Quality}

\newacronym{CELEBA}{CELEBA}{CelebFaces Attributes Dataset}

\newacronym{DDPM}{DDPM}{Denoising Diffusion Probabilistic Models}

\newacronym{E-VDVAE}{E-VDVAE}{Efficient Very Dee Variational Auto Encoder}

\newacronym{SGD}{SGD}{Stochastic Gradient Descent}

\glsenablehyper

\title{Data-Agnostic Face Image Synthesis Detection Using Bayesian CNNs
\thanks{\textit{\underline{Citation}}: 
\textbf{In progress ... }} 
}

\author{
  Roberto Leyva, Victor Sanchez, Gregory Epiphaniou,  Carsten Maple \\
  University of Warwick \\
  Coventry, UK\\
}

\begin{document}
\maketitle

\begin{abstract}
Face image synthesis detection is considerably gaining attention because of the potential negative impact on society that this type of synthetic data brings. %Hence, it is important to design robust models that can detect the synthesis process. 
In this paper, we propose a  data-agnostic solution to detect the face image synthesis process. Specifically, %our solution uses a  model that only requires real data during the training stage. Therefore, it is data-agnostic in the sense that it requires no fake face images. 
our solution is based on an anomaly detection framework that requires only real data to learn the inference process. %It uses a  model that only requires real data during the training stage. 
It is therefore data-agnostic in the sense that it requires no synthetic face images. The solution uses the posterior probability with respect to the reference data to determine if new samples are synthetic or not. Our evaluation results using different synthesizers show that our solution is very competitive against the state-of-the-art, which requires synthetic data for training.
\end{abstract}

% keywords can be removed

\section{Introduction}

Face image-based technology is fast growing for many user authentication purposes \cite{2023Maiano} making it an essential component of several authentication systems. In this context, face image synthesis poses a problem for many user profile-based systems that rely on face images, e.g., the use of fake social media accounts to spread misinformation \cite{xia2022gan,2022IAmerini} or the use of synthetic biometrics to commit identity fraud. State-of-the-art methods can generate high-quality face images with outstanding levels of featuring \cite{Guodong2019,2020VahdatArash}. Hence, it is important to accurately detect synthesized face images to reduce their negative impact on society. 

Existing solutions to detect the face image synthesis process require, unfortunately,  synthetic data at some point in the training process to learn to differentiate between real and synthetic face images. This is an important drawback because some models with undisclosed architectures can easily trick the detector by generating \emph{never-seen-before} data that looks very realistic. In this paper, we present a solution based on the anomaly detection framework, which requires training a model only with real data to learn to identify one class. This solution is then \emph{data-agnostic} in the sense that does not require any synthetic face images. Our contributions are as follows:

\begin{enumerate}
    \item We use an anomaly detection framework to detect synthetic data, which departs from the trend to use 2-class classifiers.
    \item Our proposed solution requires only real data to detect the synthesis process using a probabilistic approach.
    \item Our solution achieves very competitive performance, outperforming several state-of-the-art solutions.
\end{enumerate}

The rest of this paper is organized as follows. In Section \ref{sec:soa}, we review the most related work. In Section \ref{sec:method}, we present the proposed solution. Section \ref{sec:experiments} provides experimental results and Section \ref{sec:conclusion} concludes this paper.

\section{Related Work}\label{sec:soa}

The majority of the work related to the detection of the face image synthesis process is also related to deepfake detection. Such detection methods require detecting the faces at some point in the process, as synthetic images usually depict artifacts in the depicted faces \cite{2018Khodabakhsh,RLeyva23}. For example, Afchar \textit{et al}. \cite{2018Afchar} propose a \acrfull{CNN} based on the InceptionV3 model \cite{szegedy2016rethinking} to detect synthetic face images in videos. Their method uses the Viola-Jones face detector followed by registration, alignment, and scaling. It detects the synthesis process frame-by-frame by giving a score to each frame depicting a face. Hsu \textit{et al}. \cite{2018Hsu} propose a \acrfull{GAN}-based solution that requires measuring the contrastive loss given by the \acrshort{GAN} discriminator. Because their solution requires measuring the reconstruction error of the \acrshort{GAN}, a secondary \acrfull{SVM} is used to detect the synthesis process using the discriminator loss. Marra \textit{et al}. \cite{2018Marra} inspect a set of well-established generic models for image-related tasks, e.g. IV3, DenseNet, Xception, \cite{szegedy2016rethinking,2017Huang,2017Chollet}, to detect synthetic face images. Their work reveals that standard architectures are inherently structured to detect the synthesis process. Nataraj \textit{et al}. \cite{2019Nataraj} propose detecting synthetic face images by using a set of co-occurrence matrices prior to a \acrshort{CNN}. The authors suggest that a more descriptive input space can be generated by a set of cascade filters to detect the synthesis process.  Maiano \textit{et al}. \cite{2022irene} train several existing \acrshort{CNN} backbones to detect the synthesis process in several color spaces. Their results show that architectures are very sensitive to the color space used for detection. Rossler \textit{et al}. \cite{2019Rossler} propose to perform a series of manipulations to obtain more synthetic face images to train models. These manipulations include blending, 3D distortion, texturization, and 2D wrapping. Zhang \textit{et al}. \cite{2019Zhang} propose learning to detect the face image synthesis process by solving an image-to-image translation problem simulating artifacts. Their work, which uses a \acrshort{GAN}, shows that synthetic samples comprise low-level features visible in the Fourier domain. A further analysis of several patches is used to find distinctive patterns, thus the detection is based on spotting several artifacts. Similar spectral analyses are proposed by Frank \textit{et al}. \cite{2021Frank} by analyzing the \acrfull{DCT}. The idea is that some types of synthesis can be easily detected under a more descriptive spatial and frequency transformation. Tolosana \textit{et al}. propose to detect the face image synthesis process by means of facial landmarks \cite{2021Tolosana}. Their work suggests that separate fused models can detect the synthesis by separately analyzing several face components, e.g., the nose and eyes. This methodology is also supported by the fact that some synthesizers can only replace part of the face instead of generating a whole new face \cite{Tolosana2020}. Local and global matching is also explored by Favorskaya \textit{et al}. \cite{20121Favorskaya}; however, their method heavily relies on additional features, e.g., those extracted from the background and areas surrounding the face. Fusing models to detect the synthesis process in videos is explored by Coccomini \textit{et al}. Their method requires analyzing the faces frame by frame. It combines a \acrshort{CNN} and the recently proposed Vision Transformer \cite{2020Dosovitskiy}. Wang \textit{et al.} \cite{WangSY20} propose a \acrshort{CNN} to detect synthetic images in general. However, their work can also be used to detect synthetic face images. Other recent work \cite{Larue23} suggests adding artificially generated artifacts and then proceeding to detect the synthetic faces.   % This approach is meant to detect any sort of image, not specifically targeting faces. Yet the approach manages to detect synthetic samples. 

As discussed in this section, existing \acrshort{CNN} architectures are well-suited to detect the face image synthesis process \cite{2018Afchar,2018Marra,WangSY20,2019Nataraj}. However, they should be designed to capture the fine details of the face, which usually depict imperfections and artifacts associated with the synthesis process~\cite{2021Tolosana,Tolosana2020,20121Favorskaya}. %Some feature spaces could help to enhance the detection \cite{2019Rossler, 2022irene, 2019Zhang}. 
To this end, we design our solution using such standard \acrshort{CNN} architectures while making sure to preserve the fine details of face images. However, differently from most common solutions, we use an anomaly detection framework. %We formulate our approach preserving the feature space for only one target representation where the real data can be identified in the next section. 

% [Reviewer 1] * The authors should clearly report the references from which the formulation and derivations in Section 2 are adopted.

% [RL] As exposed in this section, existing \acrshort{CNN} architectures are well-suited to detect face synthesis \cite{2018Afchar,2018Marra,WangSY20,2019Nataraj}, yet require being designed to capture small detailing comprising imperfections \cite{2021Tolosana,Tolosana2020,20121Favorskaya} whereas some feature spaces could help to enhance detection \cite{2019Rossler, 2022irene, 2019Zhang}. We derive our approach from the aforementioned standard architectures, keeping in mind preserving the small detailing. We formulate our approach preserving the feature space for only one target representation where the real data can be identified in the next section. 

\section{The Proposed Solution}\label{sec:method}

\section{The Proposed Solution}\label{sec:method}

% [Reviewer 2] 1) It is unclear whether the approach described in Section 3 is a well-known method for solving anomaly detection tasks or a new approach.

% Although the anomaly detection framework is a well-know method, to the best of our knowledge, it has not being formulated before for the case of face synthesis detection, here is an important contribution in this work

Although strictly speaking the face image synthesis detection task is a binary classification problem aimed to determine whether a face image is real or synthetic, we assume that we have no information about the synthesizer. This is particularly useful when the attacker, who aims at synthesizing face images with malicious intentions, does not publicly disclose their model. Our proposed solution then aims at detecting synthetic face images without requiring \textbf{any} synthetic samples from any synthesizer at \textbf{any stage}. To this end, we use an anomaly detection framework. Although the anomaly detection framework is a well-known method, it has not been fully exploited for the detection of the face image synthesis process. Although the work in \cite{Larue23} also uses a one-class classifier within the context of anomaly detection, it relies on a set of local image perturbations added to real images to detect synthetic images using anomaly scores. Our work differs from that approach as it relies on a model that uses only one class with no perturbations to maximize the \acrfull{MAP}, i.e., the probability of observing the samples. In this context, samples that do not fit the positive class (normal) are deemed to be part of the negative class (abnormal) \cite{RLeyva17b}. To this end, we train a model exclusively with real face images and without the need to add any perturbations to the real data. We then use the trained model with \emph{never-seen-before} samples from both classes, i.e., real and synthetic images. Our solution uses a fine-to-coarse Bayesian \acrshort{CNN}, i.e.,  a set of convolutional layers followed by a Bayesian model implemented by \acrfull{FC} layers. Bayesian models have recently been shown to be robust to overfitting and can effectively solve problems related to sub-parametrization \cite{2022Sanae}. Because we are only modeling one class, Bayesian models are then very convenient for this task.

Formally, let us define a set of images organized as the design matrix $X = \lbrace x_1, x_2, \ldots x_N \rbrace$. Let us use define a neural network with $L$ \acrshort{FC} layers and output $y$ as follows:

\begin{equation}
    y = f^L(w^{L}, \ldots f(w^{L-1}, \ldots(f^{1}(w^{1}, z)) ) ) ,
\end{equation}

\begin{figure*}
\includegraphics[width=\textwidth]{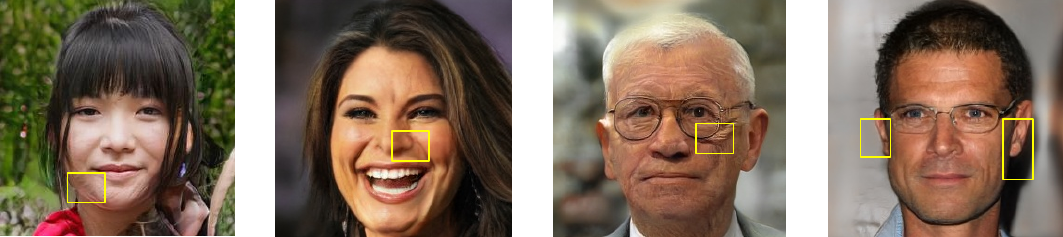}
\caption{ \small{  Sample artifacts in synthetic face images (left to right). Woman with ripples close to the chin, woman with unpaired nostril, man with scar below his left eye, and man with uneven ears.  }  }
\label{fig:artifacts}
\end{figure*}

\begin{table*}[!ht]
\centering
\vspace{0pt}
\caption{\small{ Architecture of the proposed fine-to-coarse Bayesian \acrshort{CNN}. } }
\label{table:bcnn}
\resizebox{0.7\linewidth}{!}{
\begin{tabular}{lllll}
\toprule
Layer Type & Number of Filters & Feature Map Size & Kernel Size & Stride \\
\midrule
Input &  & $256 ^{2} \times 3$ &  &  \\
Center-Crop &  & $224 ^{2} \times 3$ &  &  \\
RGB-normalization &  & $224 ^{2} \times 3$ &  &  \\
\midrule
Convolution Layer 1\textsuperscript{st} & $16$ & $224 ^{2} \times 16$ & $5 \times 5 $ & $1 \times 1$ \\
Mean Pooling Layer 1\textsuperscript{st} &  &  & $4 \times 4 $ & $2 \times 2$ \\
Sigmoid activation &&&& \\
\midrule
Convolution Layer 2\textsuperscript{nd} & $24$ & $106 ^{2} \times 24$ & $5 \times 5 $ & $1 \times 1$ \\
Mean Pooling Layer 1\textsuperscript{nd} &  &  & $4 \times 4 $ & $2 \times 2$ \\
Sigmoid activation &&&& \\
\midrule
Convolution Layer 3\textsuperscript{th} & $32$ & $50 ^{2} \times 32$ & $5 \times 5 $ & $1 \times 1$ \\
Mean Pooling Layer 1\textsuperscript{th} &  &  & $4 \times 4 $ & $2 \times 2$ \\
Sigmoid activation &&&& \\
\midrule
Batch Normalization && $ 22 ^{2} \times 32 \rightarrow 22 ^{2} \times 32$ && \\
\midrule
Fully Connected Layer 1\textsuperscript{st} &  & $ 22 ^{2} \times 32 \rightarrow 512$ &  &  \\
Dropout &&&& \\
\midrule
Fully Connected Layer 2\textsuperscript{nd} &  & $512 \rightarrow 1$ &  &  \\
\bottomrule

\end{tabular}
}
\end{table*} 
\vspace{0pt}

\noindent where $f^l(w^l, z)$ denotes the mapping function at layer $l$ with parameters $w^l$, and $z \in \mathrm{R}^{d}$ represents the latent feature space generated by a set of convolutional layers. The objective is then to train the Bayesian model that approximates $w^l$ for each \acrshort{FC} layer $l$ by using the set of probabilistic parameters, $\theta = \lbrace \alpha, \beta \rbrace$, representing the mean and variance, respectively. The output $y$ can then be modeled as the conditional Gaussian distribution $p(y \vert z)$ with inverse variance $\beta ^{-1}$:

\begin{equation}
    p(y \vert z, w, \beta) = \mathcal{N} \left( y \vert f(z,w), \beta ^{-1} \right)
\end{equation}

\noindent where $ p(w, \alpha) = \mathcal{N}(w \vert 0, \alpha ^{-1} \mathbf{I})$, with $\mathbf{I}$ as the identity matrix. For $N$ observations in $X$ with target values $\mathcal{D} = \lbrace y_1, y_2 , \ldots y_N \rbrace$, the likelihood function is:

\begin{equation}
    p(\mathcal{D} \vert w, \beta) = \prod \limits_{n=1}^{N} \mathcal{N}(y_n \vert f(z_n, w), \beta ^{-1}).
\end{equation}

The desired posterior distribution is then:

\begin{equation}
    p(w \vert \mathcal{D}, \alpha, \beta) \approx p(w \vert \alpha) ~ p(\mathcal{D}\vert w, \beta) .
\end{equation}

It can be proved \cite{bishop2006pattern} that the parameter set given by the \acrshort{MAP} is as follows:

\begin{equation}
    p(y \vert z, \mathcal{D}, w, \beta) = \mathcal{N} \left( y \vert f(z, w_{\mathrm{MAP}}), \sigma^{2}(z) \right),
\end{equation}

\noindent where the input-dependent variance $\sigma$ is given by:

\begin{subequations}
    \begin{equation}
    \sigma^{2}(z) = \beta^{-1}+g^{\top}(\alpha ~ \mathbf{I} + \beta ~ \mathbf{H}) ^{-1}g,
\end{equation}
    \begin{equation}
g = \nabla_{w} y(z ~ \vert ~ w) \big\vert _{w = w_{\mathrm{MAP}}},
   \end{equation} 
\end{subequations}

\noindent  where $\mathbf{H}$ is the Hessian matrix comprising the second derivatives of the sum of square errors with respect to the components of $w$. The distribution $p( y ~ \vert z, \mathcal{D})$ is Gaussian whose means are given by the network mapping function $f(w_{\mathrm{MAP}},z)$ and maximizes the posterior likelihood. To classify a sample $x$ as synthetic we can then use a threshold $\gamma$ on the posterior :
\begin{figure*}[t]
\begin{subfigure}{0.95\textwidth}
    \centering
    \includegraphics[width=1\textwidth]{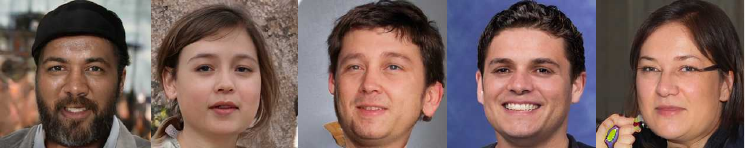}
    \caption{ \small{ XL-\acrshort{GAN} samples  } }
\end{subfigure}
\begin{subfigure}{0.95\textwidth}
    \centering
    \includegraphics[width=1\textwidth]{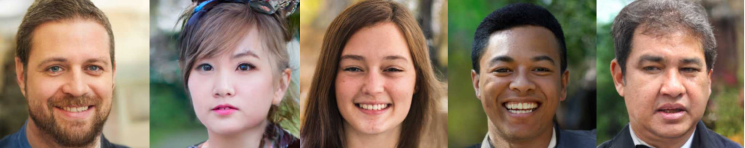}
    \caption{ \small{ \acrshort{DDPM} samples } }
\end{subfigure}\hskip 1em%
\begin{subfigure}{0.95\textwidth}
    \centering
    \includegraphics[width=1\textwidth]{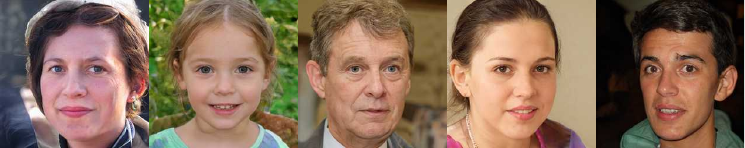}
    \caption{ \small{InsGen samples} }
\end{subfigure}\hskip 1em%

\begin{subfigure}{0.95\textwidth}
    \centering
    \includegraphics[width=1\textwidth]{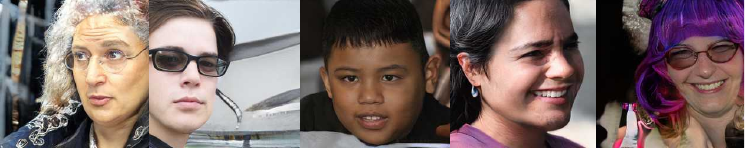}
    \caption{ \small{S\acrshort{GAN}2 samples } }
\end{subfigure}\hskip 1em%

\vspace{-5pt}
        
\caption[]{\small{Example of the synthetic face images generated by the synthesizers.}}\label{fig:synthesizers}
\end{figure*}

\begin{equation}\label{eq:thres}
    \gamma < f(w,x).
\end{equation}
under the assumption that the posterior for real images is greater than that for synthetic images:

\begin{equation}\label{eq:assumption}
   f(w,x_{\mathrm{~REAL}}) > f(w,x_{\mathrm{~FAKE}}).
\end{equation}

\noindent where $x_{\mathrm{~REAL}}$ is a real face image sample and $x_{\mathrm{~FAKE}}$ a synthetic one. Note that Eq. \ref{eq:assumption} is the foundation of the anomaly detection framework. In this work, we select the threshold $\gamma$ by inspecting the posteriors of real samples after training, which may cause the threshold to vary based on the model's initial set of learnable parameters. 

\subsection{Fine-to-coarse Bayesian CNN}

As suggested in \cite{2019Zhang},  detecting synthetic face images can be effectively performed by detecting small visual imperfections and artifacts, e.g., unexpected wrinkles, scars, and small deformations. Fig. \ref{fig:artifacts} shows several synthetic face images with visible artifacts. One can see that the synthesis process can indeed produce visible imperfections in the form of distortions or unusual human trait formations. %Thus, our fine-to-coarse Bayesian \acrshort{CNN} expands the filter banks that extract features through convolutions before feeding the extracted features to the \acrshort{FC} layers. 
Because we are interested in expanding the spatial information extracted from the images, our fine-to-coarse Bayesian \acrshort{CNN}  progressively increases the number of filters along the convolutions layers before feeding the extracted features to the \acrshort{FC} layers. Furthermore, to minimize the information loss in the pooling stages,  we employ mean pooling operations to reduce the loss 
 of important visual details, especially 
 the artifacts in synthetic face images, which tend to be quite small. %
%\hl{Since pooling and large stride operations may hinder performance by condensing spatial information.} Because the artifacts in fake face images are usually quite small, these operations can easily discard important visual details.  
Table \ref{table:bcnn} summarizes the architecture of the proposed fine-to-coarse Bayesian \acrshort{CNN}. Note that the two \acrshort{FC} layers form a \acrfull{MLP} structure as the decision layers and constitute the Bayesian model. To produce large positive output values, we employ the Sigmoid activation function for all feature maps. Thus, the \acrshort{MLP} receives only positive values. %Our experiments show that adding dropout in the \acrshort{FC} layers produces no visible improvements, hence, we do not use dropout.

\section{Experiments}\label{sec:experiments}

We perform experiments using the face image datasets \acrfull{FFHQ} \footnote{\url{https://github.com/NVlabs/ffhq-dataset}} and \acrfull{CELEBA} \footnote{\url{https://github.com/tkarras/progressive_growing_of_gans}}  \cite{2017KarrasTero,2015LiuZiwei}, which comprise 70K and 30K real face image samples, respectively. Let us recall that our solution only requires real samples for training. However, to evaluate performance in detecting synthetic face images, we use four synthesizers to generate several synthetic face images. Specifically, we use the pre-trained models provided by the authors of these four synthesizers:  S\acrshort{GAN}2 \cite{2020KarrasTero}, XL-\acrshort{GAN} \cite{2022Sauer}, InsGen \cite{2021Yang}, and \acrfull{DDPM} \cite{2020HoJonathan} \footnote{\url{https://github.com/hojonathanho/diffusion}}. Fig. \ref{fig:synthesizers} shows several samples generated by these four synthesizers. To have synthetic samples for evaluation along with the real samples in the  \acrshort{FFHQ}  dataset, we generate 224K synthetic images, 56K generated by each of the four synthesizers. All 224K synthetic images are the same size as the real images in the  \acrshort{FFHQ}  dataset and are in an uncompressed format. For the case of the \acrshort{CELEBA} dataset, we generate 72K synthetic images to be used for evaluation along with the real samples, 24K synthetic images generated by each of the four synthesizers. All 72K synthetic images are the same size as the real images in the  \acrshort{CELEBA}  dataset and are in an uncompressed format. 

%To evaluate our approach, we use four different synthesizers to generate $56k+56k+56k+56k = 224k$ from the \acrshort{FFHQ}  dataset, i.e., $56k$ from each synthesizer, and $24k+24k+24k = 72k$ samples from the \acrshort{CELEBA} dataset. The real and synthetic images have the same size %$256^2$ 
%in an uncompressed format. The synthesizers used are S\acrshort{GAN}2 \cite{2020KarrasTero}, XL-\acrshort{GAN} \cite{2022Sauer}, InsGen \cite{2021Yang}, and \acrfull{DDPM} \cite{2020HoJonathan} \footnote{\url{https://github.com/hojonathanho/diffusion}} using the pr-trained synthesizer models publicly available. For all the synthesizers we use the pre-trained models the authors provide. 
%\\

\begin{table*}[!t]
\centering
\caption{\small{ \acrshort{mAp} values ($\uparrow$) of several solutions for different synthesizers and split values for the \acrshort{FFHQ} dataset. The best (second best) results are highlighted in \textbf{bold} (\underline{underlined}).} }
\label{table:resultsFFHQ}
\resizebox{0.5\linewidth}{!}{
\begin{tabular}{llcccc}
\toprule
\multirow{2}{*}{Method} & \multirow{2}{*}{Synthesizer} & \multicolumn{4}{c}{ Split (\% of data used for training )} \\
&& 20\% & 40\% & 60\% & 80\% \\
\midrule
\multirow{4}{*}{ \acrshort{DCT}-Ridge \cite{2021Frank} } 
& S\acrshort{GAN}2 \cite{2020KarrasTero}    & 0.492 & 0.533 & 0.654 & 0.761 \\
& InsGen \cite{2021Yang}                    & 0.501 & 0.534 & 0.583 & 0.741 \\
& \acrshort{DDPM} \cite{2020HoJonathan}     & 0.505 & 0.512 & 0.559 & 0.721 \\ 
& XL-\acrshort{GAN} \cite{2022Sauer}        & 0.511 & 0.522 & 0.544 & 0.698 \\

\multirow{4}{*}{ DF \cite{2022irene} } 
& S\acrshort{GAN}2      & 0.503 & \underline{0.575} & \underline{0.701} & \underline{0.816} \\
& InsGen                & \underline{0.551} & \underline{0.584} & \textbf{0.731} & \textbf{0.802} \\
& XL-\acrshort{GAN}     & 0.565 & 0.566 & \underline{0.624} & \underline{0.732} \\
& \acrshort{DDPM}       & 0.518 & 0.563 & 0.691 & 0.791 \\   

\multirow{4}{*}{ Auto\acrshort{GAN} \cite{2019Zhang} } 

& S\acrshort{GAN}2 & \underline{0.513} & 0.544 & 0.603 & 0.729 \\
& InsGen  & 0.544 & 0.576 & 0.623 & \underline{0.787} \\
& \acrshort{DDPM} & 0.512 & 0.525 & 0.557 & 0.653 \\   
& XL-\acrshort{GAN} & 0.504 & 0.518 & 0.544 & 0.642 \\
  
  \midrule

\multirow{4}{*}{ Bayesian\acrshort{CNN} } 
& S\acrshort{GAN}2  & \textbf{0.629} & \textbf{0.683} & \textbf{0.754} & \textbf{0.843} \\
& InsGen  & \textbf{0.552} & \textbf{0.593} & 0.643 & 0.771 \\
& \acrshort{DDPM} & 0.562 & 0.595 & 0.667 & \textbf{0.783} \\ 
& XL-\acrshort{GAN}  & \textbf{0.573} & \textbf{0.597} & \textbf{0.643} & \textbf{0.793} \\

\bottomrule
\\

\end{tabular}
}
\end{table*}

\begin{table*}[!t]
\centering
\caption{\small{ \acrshort{mAp} values ($\uparrow$) of several solutions for different synthesizers and split values for the \acrshort{CELEBA} dataset. The best (second best) results are highlighted in \textbf{bold} (\underline{underlined}).} }
\label{table:resultsCELEB}
\resizebox{0.5\linewidth}{!}{
\begin{tabular}{llcccc}
\toprule
\multirow{2}{*}{Method} & \multirow{2}{*}{Synthesizer} & \multicolumn{4}{c}{ Split (\% of data used for training )} \\
&& 20\% & 40\% & 60\% & 80\% \\
\midrule
\multirow{3}{*}{ \acrshort{DCT}-Ridge \cite{2021Frank} } 
& S\acrshort{GAN}2 \cite{2020KarrasTero}    & 0.562 & 0.593 & 0.681 & 0.813 \\
& \acrshort{DDPM} \cite{2020HoJonathan}     & 0.578 & 0.594 & 0.615 & 0.794 \\ 
& XL-\acrshort{GAN} \cite{2022Sauer}        & 0.566 & 0.573 & 0.602 & 0.778 \\

\multirow{3}{*}{ DF \cite{2022irene} } 
& S\acrshort{GAN}2      & 0.552 & 0.642 & 0.770 & 0.833 \\
& \acrshort{DDPM}       & 0.575 & 0.654 & 0.723 & 0.805 \\   
& XL-\acrshort{GAN}     & 0.602 & 0.614 & 0.693 & 0.791 \\

\multirow{3}{*}{ Auto\acrshort{GAN} \cite{2019Zhang} } 

& S\acrshort{GAN}2 & 0.562 & 0.612 & 0.669 & 0.802 \\
& \acrshort{DDPM} & 0.570 & 0.593 & 0.653 & 0.733 \\   
& XL-\acrshort{GAN} & 0.562 & 0.587 & 0.644 & 0.702 \\
  
  \midrule

\multirow{3}{*}{ Bayesian\acrshort{CNN} } 
& S\acrshort{GAN}2  & 0.664 & 0.743 & 0.773 & 0.843 \\
& \acrshort{DDPM} & 0.602 & 0.655 & 0.694 & 0.796 \\ 
& XL-\acrshort{GAN}  & 0.632 & 0.667 & 0.730 & 0.812 \\

\bottomrule
\\

\end{tabular}
}
\end{table*}

% [Reviewer 1] * Three methods are used for comparison. One of them is actually proposed for the detection of deepfake dynamic videos, which are significantly different from GAN-generated static images. The comparison should include specific methodologies for synthetic images, such as the one in: D. Gragnaniello, D. Cozzolino, F. Marra, G. Poggi, L. Verdoliva, "Are GAN generated images easy to detect? A critical analysis of the state-of-the-art", IEEE ICME 2021 The code is available at: https://github.com/grip-unina/GANimageDetection

% [RL] To make comparisons with existing methods we use a similar strategy as suggested by Gragnaniello \textit{et al.} \cite{Gragnaniello21}. However, we are interested only in the capacity of detecting faces synthesis. This is it: we do not evaluate the synthesizers across diverse dataset to learn the synthesis process but only faces datasets to evaluate the capacity of detection using unseen samples.

%\textbf{Training parameters}: %We train our fine-to-course Bayesian \acrshort{CNN}  using an error plateau detector, scaling the learning rate by a factor of ten. We use the \acrfull{ELBO} criterion during training for the Bayesian model. 
% \hl{CAN YOU CONFIRM THIS IS BETWEEN THE TEST AND VALIDATIONS SETS AND NOT BETWEEN THE TRAINING AND VALIDATION SETS?} Yes it is

Our fine-to-course Bayesian \acrshort{CNN} is implemented in \textit{pyro} \footnote{\url{https://pyro.ai/}} using two GTX 1080 TI GPUs. We use an exponential learning rate scheduler having \acrfull{SGD} as the backbone starting at $10^{-3}$ with a decay factor of $0.1$. We use a TraceGraph \acrfull{ELBO} loss function as a back-propagator and monitor the loss plateau on the validation and training sets. Initially, we use $50$ epochs and when the model achieves a 1\% improvement in accuracy with respect to the previous validation iteration, we use it as the best model and continue iterating. Thus at the end of the training process, the best model is the one that achieves the best accuracy on the validation set. To prevent overfitting, we have an early stop criterion of $6\%$ %$0.06$ 
between the accuracy achieved on the test set and the accuracy achieved on the validation set. The convolution banks are preset with Xavier initialization. We use batches of $5122$ samples.

%--------------------------------------------------
\begin{figure*}[!ht]
\centering
\includegraphics[width=0.8\linewidth]{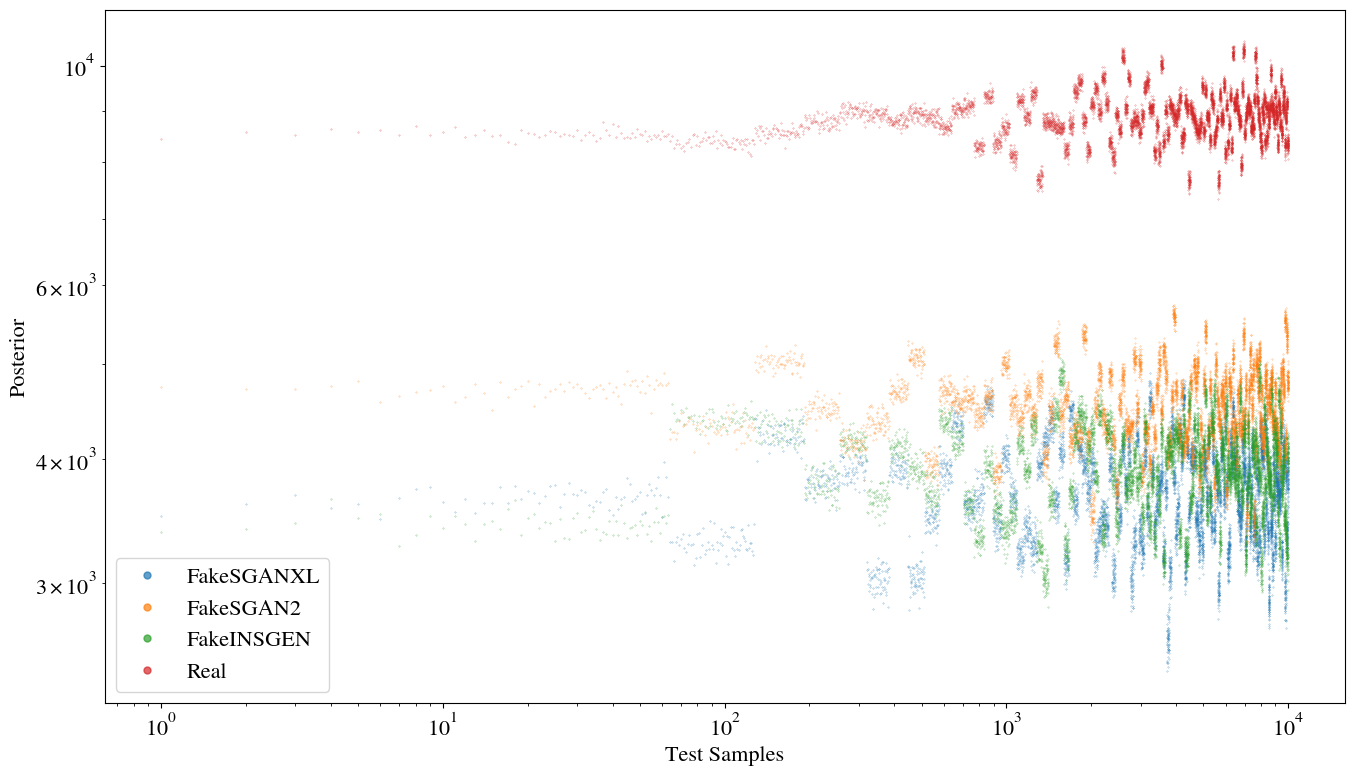}
\caption{  \small{ Posterior values produced by the proposed Bayesian \acrshort{CNN}. The real and synthetic samples form two distinct regions. Thus, we can set the posterior threshold accordingly. }}
\label{fig:posterior}
\end{figure*}
%--------------------------------------------------
\begin{figure*}[t]
\begin{subfigure}{0.3\textwidth}
    \centering
    \includegraphics[width=1\textwidth]{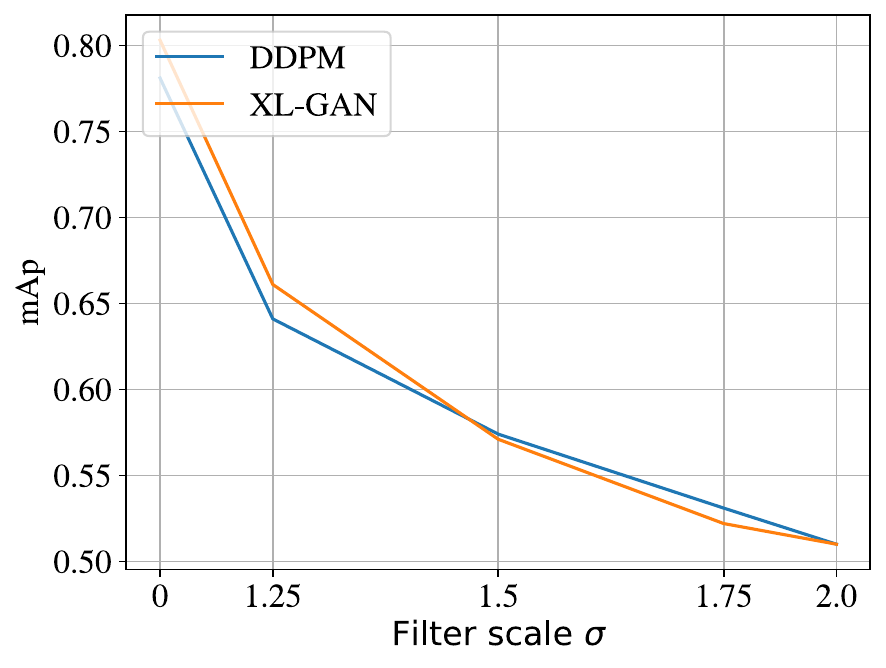}
    \caption{ \small{ Blurring.  } }\label{fig:blur}
\end{subfigure}
\begin{subfigure}{0.3\textwidth}
    \centering
    \includegraphics[width=1\textwidth]{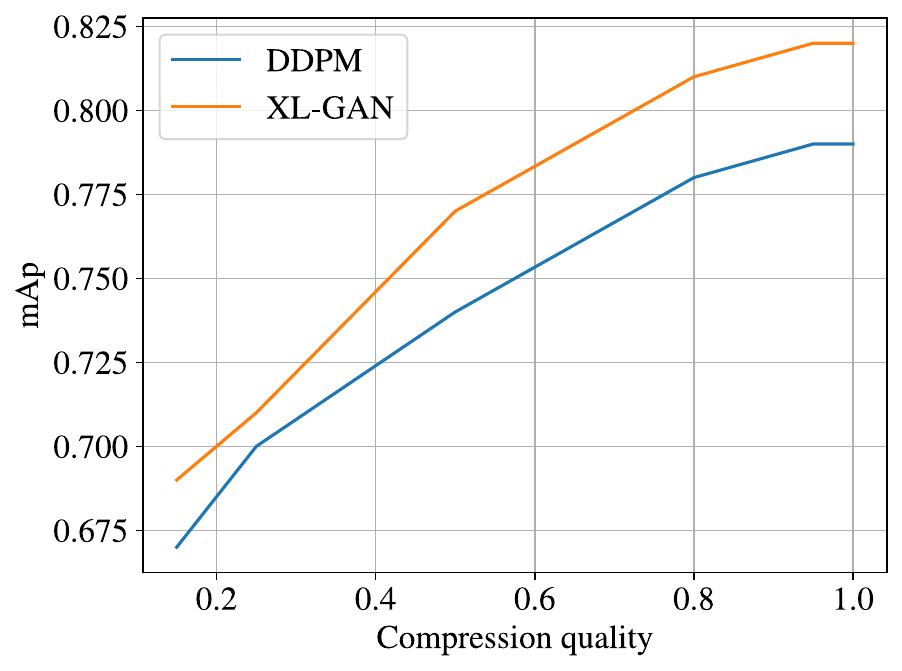}
    \caption{ \small{JPEG compression. } }\label{fig:compress}
\end{subfigure}\hskip 1em%
\begin{subfigure}{0.3\textwidth}
    \centering
    \includegraphics[width=1\textwidth]{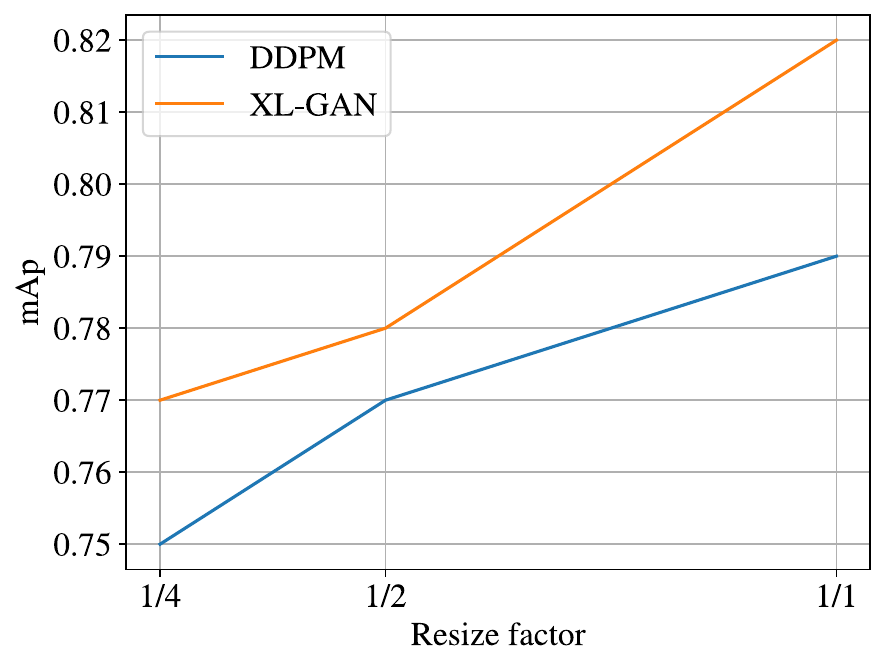}
    \caption{ \small{ Resizing. } }\label{fig:resize}
\end{subfigure}\hskip 1em%
        
\caption[]{\small{Performance or the proposed solution when post-processing is used on the test images. }}\label{fig:proceff}
\end{figure*}

% here explain the part of the single class setup 

% what happens on the splits
% train as one class cassifier
% what strategy was Grangniello??
% we use a similar strategy \hl{WHAT IS THIS STRATEGY? PLEASE SPECIFY IT HERE AS THE READER MAY NOT KNOW WHAT STRATEGY YOU ARE TALKING ABOUT}

% To Victor: All the other methods see the synthetic FOR TRAINING.

% "Specifically, we train the other methods with the samples generated by one synthesizer and test them with the remaining unseen data, which includes the samples generated by the other unseen synthesizers. They never saw samples from that synthesizer before for training."

% We then use the unseen data for testing, which includes $50\%$ real unseen data + $50\%$ synthetic data. That 50% is coming from the syntheziser which the method HAS NEVER SEEN SAMPLES BEFORE

To make comparisons with existing methods, we use a similar strategy as that suggested by Gragnaniello \textit{et al.} \cite{Gragnaniello21}, which is a strategy for synthetic images in general, not exclusively face images. Their strategy requires training on a reference dataset targeting one class out of ten and testing on different image scales. % for all the compared methods. \
Their strategy uses %as a synthetic source 
seven synthesizers to generate around 39K synthetic samples in an %from all following an 
imbalanced fashion; i.e., more samples from some synthesizers than others. In this work, we are interested only in evaluating the capacity to detect synthetic face images regardless of the image scale. We then focus on evaluating the detection of %, we only use faces as objects to evaluate the capacity of detection using 
unseen samples at one scale with balanced data generated by four synthesizers. \\

We compare our solution against the methods proposed in \cite{2021Frank, 2019Zhang, 2022irene}. %We train these methods using the same real data used for training our solution plus some synthetic data generated by the four synthesizers. %different synthetic sources balancing the number of samples per synthesizer (i.e. all synthesizers generate the same number of samples). 
 %Specifically, we train these methods with the samples generated by one synthesizer and test them with the remaining unseen data, which includes the samples generated by the other synthesizers. 
 These methods are trained to detect real samples as the class $1$ and the synthetic samples as the class $0$. Specifically, we train these methods %To compare against the work in \cite{2021Frank}, 
 with a proportion of the real samples defined by the split used plus the same number of synthetic samples generated by one of the four synthesizers. %, which we call the reference synthesizer.  %S\acrshort{GAN}2 to train their ridge regression. 
 We then use unseen data for testing, which includes the same proportion of unseen real samples and unseen synthetic samples. We repeat this process with both datasets and the other synthesizers.  To compare against the method in \cite{2022irene}, we only use the RGB color space. %three times as the accuracy for the reference synthesizer is above 99\%, as reported in \cite{2021Frank}. To compare against the work in \cite{2019Zhang}, we augment the training dataset produced by the S\acrshort{GAN}2 and test with the unseen data excluding the data produced by the reference synthesizer. We repeat this process three times as well. To compare against the method in \cite{2022irene}, we use the samples produced by the synthesizer S\acrshort{GAN}2 for training and test with the unseen data.  We repeat this process three times and only use the RGB color space. 

For the method in \cite{2019Zhang} \footnote{\url{https://github.com/ColumbiaDVMM/AutoGAN}}, we keep all the default settings from the implementation and only append the tree structure of the real/synthetic faces. No threshold is set to detect synthetic face images but only the output of the discriminator. For the method in \cite{2022irene}, we train from zero a model using the reported parameters and set the classification threshold at $0.7$ from the last decision layer as it is not specified by the authors. We also add Sigmoid activations as the authors report the use of a binary cross entropy loss. For the method in \cite{2021Frank}, we employ a grid search to find the best parameters as the authors report for the described \acrshort{CNN}. We set a classification threshold at $0.9$ that empirically provides good results. For our solution, we maximize the \acrshort{MAP} until a plateau is observed. We set the threshold $\gamma$ in Eq. \ref{eq:thres} after inspecting a few samples from the posterior distribution. In this case, the test samples are deemed real/synthetic after manually inspecting the validation set. Because the means and variances of the model are randomly initialized, we observe that the threshold should change for every run. The reported results in Tables \ref{table:resultsFFHQ} and  \ref{table:resultsCELEB} then use a different threshold for each split.

% \hl{I THINK THE PROBLEM HERE IS THAT IT IS NOT CLEAR HOW A SPLIT OF 20\% FOR A METHOD THAT IS NOT BASED ON ANOMALY DETECTION COMPARES WITH THE SAME SPLIT FOR OUR SOLUTION. OUR SOLUTION USES ONLY NORMAL DATA, HENCE A 20\% SPLIT INCLUDES ONLY NORMAL DATA BUT FOR ANY OTHER METHOD IN THIS TABLE, THIS 20\% SPLIT SHOULD INCLUDE FAKE AND REAL IMAGES, HOW MUCH OF THESE 20\% IS REAL, AND HOW MUCH IS FAKE AND HERE THE FAKE IMAGES COME FROM? IF YOU CLARIFY THIS FOR EACH EVALUATED METHOD, THE REVIEWER WILL THEN UNDERSTAND THESE SPLITS}

% If you select 20% of the real data for training you have 80% for testing. If you have 100 real samples to test then you need 100 synehtic images to test too. Thus during test times we always have 50/50% real/synthetic samples to evaluate. Thats why we have 56K synthetic samples per synthesizer, to always balance the remaining real unseen samples.

Table \ref{table:resultsFFHQ} and  \ref{table:resultsCELEB} tabulate results for the real images of the \acrshort{FFHQ} dataset and the \acrshort{CELEBA} dataset, respectively, in terms of the \acrfull{mAp} values for different proportions (splits) of training data. In both tables, the tabulated splits indicate the proportion of real samples from each dataset used for training our solution.  For the case of the other evaluated methods, the tabulated splits indicate the proportion of real samples from each dataset used for training plus the same amount of training synthetic samples generated by the synthesizer tabulated in each row.  From Table \ref{table:resultsFFHQ}, we can see that the proposed solution (Bayesian\acrshort{CNN}) achieves very competitive performance when trained on the real images of the \acrshort{FFHQ}  dataset. Particularly, using 80\% of the available training data gives the best \acrshort{mAp} values for two of the synthesizers. %\hl{We select the threshold depending on the method implementation and for our method case after inspecting the \acrshort{MAP}.
%Table \ref{table:resultsCELEB} tabulates results for the real images of the \acrshort{CELEBA}  dataset in terms of the \acrfull{mAp} values for different proportions (splits) of training data. 
One can also see in Table  \ref{table:resultsCELEB} that the proposed solution also achieves very competitive performance when trained on the real images of the \acrshort{CELEBA}  dataset. Namely, our solution gives the best performance for the detection of synthetic images generated by the XL-\acrshort{GAN} and S\acrshort{GAN}2 synthesizers.

We also examine the posteriors of the data generated by each synthesizer and plot them along with the posteriors of the real data in Fig. \ref{fig:posterior}. This plot shows that it is indeed possible to distinguish the synthetic samples from the real ones by thresholding the posterior linearly. Hence, the threshold selection in Eq. \ref{eq:thres} is appropriate as this establishes a linear margin. As we can see from this figure, the synthetic data is concentrated in a region where low posterior values exist. This further confirms that using an anomaly detection framework is an effective solution to detect synthetic face images.  Moreover, such posterior values are intrinsic to our Bayesian \acrshort{CNN}, which is expected to produce high posterior values for data that is very similar to the one used during training (i.e., real face images) and low values for \emph{never-seen} data (i.e., synthetic face images). It is important to recall that the location of the region where the synthetic samples lie varies depending on the initialization of the model's parameters. 

% [Reviewer 1] * One important issue to be evaluated is the robustness to possible post processing in the test images, such as resizing and compression, which commonly arise in the lifecycle of digital images. The authors should experimentally explore this aspect.

We also evaluate performance after applying common post-processing on the test images: (1) Blurring by varying the size of the filter scale $\sigma$; (2) JPEG compression at different qualities; and (3) resizing by a factor of $1/2$ and $1/4$ using bilinear interpolation. Fig. \ref{fig:proceff} shows the results of this experiment. Fig. \ref{fig:blur} shows that blurring has a very negative effect on performance, to the point of almost random classification for large values of $\sigma$. Fig. \ref{fig:compress} shows that very aggressive compression hinders performance, yet the effect is not as severe as the one introduced by blurring the images. Finally, Fig. \ref{fig:resize} has also a drastic effect, similar to blurring, as losing spatial information hinders the model's performance in detecting the synthetic samples. This experiment reveals that the proposed solution is very sensitive to losing the fine details of the images as our Bayesian \acrshort{CNN} relies on detecting such small artifacts and imperfections. %on the synthesis, thus spatial transformations involving information loss has negative effects on the performance.

% 3) [Reviewer 2] Figure 3 is not easy to understand, while more details on the architectural choices of the CNN would contribute to a better understanding of the choices made.

Finally, we also discuss several architectural decisions that led to the final architecture of our fine-to-coarse Bayesian \acrshort{CNN}. We observe that small kernel sizes for the convolutional layers significantly improve the performance, e.g. $3-4\%$ on the large splits, while more than three filter banks have little effect on the performance but a severe impact on processing times. Compared to using filter banks of the same size, the proposed fine-to-coarse filter bank provides $5\%$ improvement on the large splits. We observe that more than two \acrshort{FC} layers provide no significant improvement. Adding batch normalization provides faster convergence and less sensitivity to initialization. Finally, we observe that using dropouts, high posteriors can be achieved with significantly fewer parameters. Because our ultimate goal is to maximize the posterior for the real data with the fewest parameters possible, dropout is used. The proposed architecture in Table \ref{table:bcnn} then fairly trades performance for complexity.

%The last experiment is an ablation study to evaluate the effects on the posterior as the number of parameters changes and dropout is used in last \acrshort{FC} layers. Specifically, we increase the number of parameters by increasing the number of filters and modifying the kernel size starting from the parameters listed in Table \ref{table:bcnn}. Fig. \ref{fig:parms} shows the results of the experiment. One can see that by using dropout,  high posteriors can be achieved with significant fewer parameters.  Our ultimate goal is to maximize the posterior for the real data with the fewer parameters possible. Then, the proposed architecture in Table \ref{table:bcnn} trades performance for complexity fairly.

% [Reviewer 1] * While multiple synthesisers are considered, real data used for validating the approach come from one dataset only. I suggest to include in the testing set also real data which does not belong to the reference dataset (FFHQ), and assess the performance of the proposed methodology in this scenario.

% [RL] as part of future work. Such experimentation is very extensive, training on several real face datasets validating on the other and also check again the performance on synthetic dataset. I added it as future work "and conducting cross-data experiments on more real/synthetic datasets. "

\section{Conclusion}\label{sec:conclusion}

In this paper, we have proposed a solution based on anomaly detection to detect synthetic face images, which implies training using only one class. Our solution is then data-agnostic as it requires no synthetic samples during training. This is a powerful advantage as we may not have information about the synthesizer or any of the synthetic face images. For detection, the solution uses a Bayesian \acrshort{CNN}  that extracts spatial features from the face images while preserving the small details associated with common artifacts and imperfections found in synthetic face images. Our performance evaluation results show that the proposed solution can achieve very competitive accuracy, outperforming several state-of-the-art methods that require training on real and synthetic face images. Our future work focuses on improving detection accuracy especially when post-processing is used on the test images,  defining an automatic margin selection process to set thresholds, and conducting cross-data validations on more real/synthetic datasets.

%Bibliography
\bibliographystyle{unsrt}  
\bibliography{references}

\end{document}